\documentclass[letterpaper, 10 pt, conference]{IEEEtran}  % Comment this line out if you need a4paper

\IEEEoverridecommandlockouts                              % This command is only needed if 
\usepackage{graphics} % for pdf, bitmapped graphics files
\usepackage{epsfig} % for postscript graphics files
\usepackage{mathptmx} % assumes new font selection scheme installed
\usepackage{times} % assumes new font selection scheme installed
\usepackage{amsmath} % assumes amsmath package installed
\usepackage{amssymb}  % assumes amsmath package installed

\usepackage[T1]{fontenc}
\usepackage{amsfonts}
\usepackage{booktabs}
\usepackage{siunitx}
\usepackage{caption} 
\usepackage{booktabs}
\usepackage{graphicx}
\usepackage{algorithm, algpseudocode}
\usepackage{float}
\usepackage[percent]{overpic}
\usepackage{caption}
\DeclareGraphicsExtensions{.pdf,.png}

\usepackage[font=small,skip=0pt]{subcaption}

\usepackage{array, tabularx, makecell, booktabs}%
\usepackage{tikz}

\usepackage{geometry}
\usepackage{comment}

\usepackage{hyperref}
\hypersetup{
    colorlinks=true,
    linktoc=all,
    citecolor=black,
    filecolor=black,
    linkcolor=blue,
    urlcolor=black
}

\title{\LARGE \bf
Arena-Rosnav 2.0: A Development and Benchmarking Platform for Robot Navigation in Highly Dynamic Environments
}

\author{Linh K{\"a}stner$^{1,*}$\thanks{$^{1}$The Chair of Industry Grade Networks and Clouds, Technical University Berlin (TUB), Germany
		{\tt\small linhdoan@tu-berlin.de}}, Reyk Carstens$^{1,*}$, Huajian Zeng$^{2}$\thanks{$^{2}$The Chair of Robotics, Artificial Intelligence and Real-time Systems, Technical University Munich (TUM), Germany}, Jacek Kmiecik$^{1}$,\\ Teham Bhuiyan$^{1}$, Niloufar Khorsandhi$^{1}$, Volodymyir Shcherbyna$^{1}$ and Jens Lambrecht$^{1}$
		\thanks{*Contributed equally}
}

\begin{document}

\maketitle
\thispagestyle{empty}
\pagestyle{empty}

% !TeX encoding = utf-8
% !TeX language = en_GB
% !TeX spellcheck = en_GB
% !TeX root = paper.tex

\begin{abstract}

Following up on our previous works, in this paper, we present Arena-Rosnav 2.0 an extension to our previous works Arena-Bench \cite{kastner2022arena-bench} and Arena-Rosnav \cite{kastner2021towards}, which adds a variety of additional modules for developing and benchmarking robotic navigation approaches. The platform is fundamentally restructured and provides unified APIs to add additional functionalities such as planning algorithms, simulators, or evaluation functionalities. We have included more realistic simulation and pedestrian behavior and provide a profound documentation to lower the entry barrier. We evaluated our system by first, conducting a user study in which we asked experienced researchers as well as new practitioners and students to test our system. The feedback was mostly positive and a high number of participants are utilizing our system for other research endeavors. Finally, we demonstrate the feasibility of our system by integrating two new simulators and a variety of state of the art navigation approaches and benchmark them against one another. The platform is openly available at \href{https://github.com/Arena-Rosnav}{\color{blue}https://github.com/Arena-Rosnav}.

%In recent years, Deep Reinforcement learning has made remarkable progress in all kinds of application areas such as robotic control, simulation, and natural language processing. in robot navigation as well. various research works applied drl to navigate robots autonomously in unknown environments. 
%The ability to autonomously navigate safely, espe-
%cially within dynamic environments, is paramount for mobile
%robotics. In recent years, DRL approaches have shown superior
%performance in dynamic obstacle avoidance. However, these
%learning-based approaches are often developed in specially de-
%signed simulation environments and are hard to test against con-
%ventional planning approaches. Furthermore, the integration and
%deployment of these approaches into real robotic platforms are
%not yet completely solved. 

%however, most research works either focus on provideing an end to end solution training the whole system using drl or focus on one specific aspect such as local planning due to constraints such as myopic nature. in this paper, we propose another type of training called the holicstic training approach in which the training is conducted in an end to end manner. however, different from other research works, the global planner is just included as an already working entity and not training but ultiized as an utiliy to train the local planner. this way, the local planner can better adapt to choices made by the gloabl planner. We tested the proposed system against two baselines and found that the resulting navigation system could deliver more smooth navigation. 

\end{abstract}
\section{Introduction}
\noindent As human-robot-collaboration and interaction is becoming increasingly essential for tasks such as healthcare, logistics, or delivery, robots need to navigate in collaborative and highly dynamic environments. In recent years, Deep Reinforcement Learning (DRL) for navigation in dynamic environments has achieved remarkable results and was applied by a variety of researchers. However, several works have pointed out downsides and challenges when working and developing DRL approaches \cite{xiao2022motion} \cite{kastner2022arena-bench}. In particular, development is still a high hurdle to overcome, the training process is oftentimes difficult and tedious, and comparability with other approaches is not trivial. A number of works aspired to provide benchmarks to compare DRL approaches. However, their functionality is limited, the setup and usage is difficult, or they are not easily extensible or customizable. Furthermore, most benchmarks aimed to compare approaches of the same paradigm domain and lack the possibility to compare across the different paradigms.
\begin{figure}[!h]
    \centering
    \includegraphics[width=0.99\linewidth]{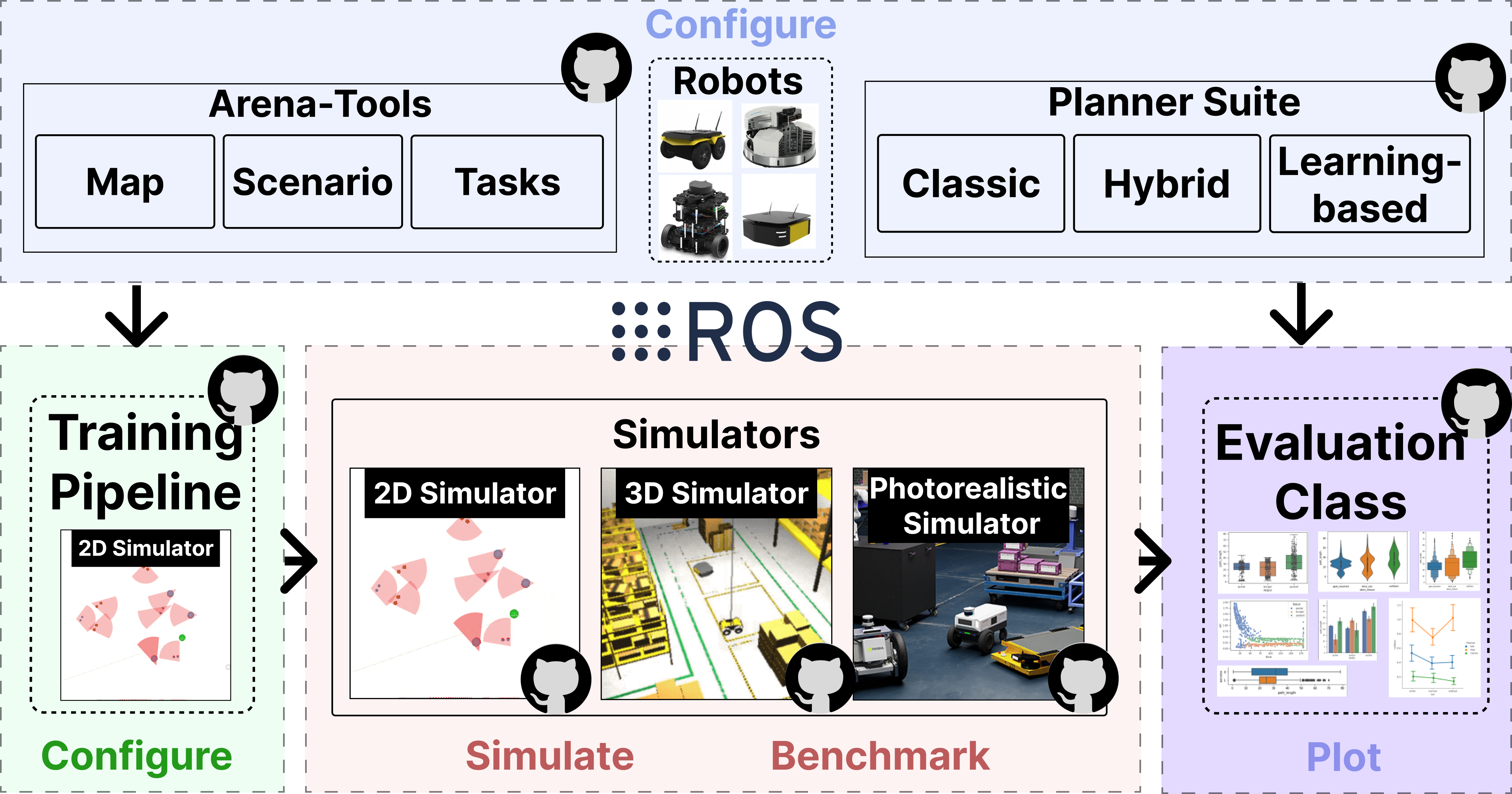}
    \caption{Arena-Rosnav 2.0 provides tools to develop, train, and benchmark DRL approaches against state-of the art navigation planners in highly dynamic and crowded environments. In contrast to the previous version, the structure of this version is completely modular with each entity being independently deployable within its own Github repository. This ensures a simplified extension of new modules such as new planners, simulation environments, or evaluation functionalities. In this work, we integrated two more simulators and six more navigation planners into our system using the provided APIs. Additionally, we provide a profound documentation and step-by-step tutorials of integrating new modules.}
    \label{intro}
\end{figure}
\noindent
In our previous works, we proposed Arena-Bench \cite{kastner2022arena-bench} to address these issues and provide an unified benchmark, which is able to compare both paradigms side-by-side. Despite a variety of functions and planners, the system had a number of limitations such as the limitation to 2D training and simulation environments, a tedious installation process, and lack of a profound documentation and tutorials which could make it hard for beginners and new practitioners to use. On that account, in this paper, we propose Arena-Rosnav 2.0 to tackle aforementioned problems and provide a more extensive development and benchmarking suite containing the most relevant functionalities to develop train and compare DRL agents for navigation in realistic dynamic environments.
Arena-Rosnav 2.0 is designed fundamentally differently than its previous version and aims to lower the barrier of entrance by simplifying setup and installation processes as well as simplifying functional extensions. Therefore, the majority of modules are independent from one another and reside in independent Github repositories. We provide APIs for our main entities to integrate new aspects and functionalities. In the process, we integrated two more simulators - a photorealisitc simulation based on Unity, and another resource-efficient 2D simulator for training. Additionally, the planning approaches are outsourced as separate modules and we integrated a high number of state-of-the-art planning approaches to be used with our system. Additionally, the movement and behavior patterns of dynamic pedestrians are now more realistic due to integration of several social navigation modules from different researchers.
The installation process is significantly simplified and the repository restructured. Thus, a more than five-fold compression could be achieved. We provide a profound documentation with step-by-step tutorials and examples to make the barrier of entrance even lower. To evaluate our system, we conducted a user study asking a number of different robotic researchers to test and validate our system. The feedback was mostly positive and some of the researchers pointed out that they are using our platform for current research endeavors. The main contributions of this work are the following:
\begin{itemize}
    \item Extension of our previous platforms with more planners, simulators, task modes
    \item Restructuring and reworking the old infrastructure to provide a more resource-efficient, scaleable, and extensible system for developing training and benchmarking navigation approaches
    \item Evaluation of our system by first, conducting a user study to validate the added value and improved user-experience and -understanding, and second, by benchmarking all planners on highly dynamic scenarios.
\end{itemize}

% \newgeometry{top=0.75in,bottom=0.75in,right=0.75in,left=0.75in}
\section{Related Works}
\noindent
This work extends our previous works Arena-Bench \cite{kastner2022arena-bench} and Arena-Rosnav \cite{kastner2021towards}, where a platform for developing and training DRL agents within a resource-efficient 2D environment was proposed and tools were provided to benchmark the trained agent against several model-based and learning based approaches on different highly dynamic scenarios respectively.
\\\noindent
DRL based approaches have shown remarkable results for navigation in highly dynamic environments and a variety of research works incorporated DRL into their systems \cite{dugas2020navrep}, \cite{chen2019crowd}, \cite{chen2017socially} \cite{faust2018prm}, \cite{kastner2021arena}, \cite{kastner-aio}. However a major problem of DRL approaches is the lack of an unified benchmark to validate the approaches against conventional navigation systems \cite{xiao2022motion}. As aspired by a number of researchers benchmarks to compare state-of-the-art planners were presented in recent years. 
\\\noindent
A benchmark suite named Bench-mr for motion planning approaches was presented by Heiden et al. \cite{bench_mr}. It provides several modules to test, evaluate, and compare different motion-planning techniques in static environments. Robobench by Weisz et al. \cite{robobench} offers a range of robotic tasks that include manipulation and navigation. Other benchmarks that focus on navigation in static environments are presented in the works of Althoff et al. \cite{althoff2017commonroad} or Moll et al. \cite{moll2015benchmarking}. 
\\\noindent
However, with the rising necessity for human-robot-collaboration, navigation not only in static but also in dynamic environments become much more essential. Thus, in recent years, a variety of platforms were proposed that enable training and testing navigation approaches in crowded and collaborative environments \cite{everett2018motion}, \cite{chen2019crowd}. 
\\\noindent
Tsoi et al. \cite{tsoi2020sean} presented SEAN a benchmark for social navigation, which provides tools to develop navigation planners for collaborative environments with humans. They extended their works in SEAN 2.0 \cite{tsoi2022sean} were they included more realistic movement behavior and more robots. Their simulator contains realistic 3D simulations and is build on top of Unity. however, there are limitations to the platform as only a specific number of worlds and robots can be selected. Other benchmarks that focus on navigation in social situations include Socialgym by Holtz et al. \cite{holtz2021socialgym} or HuNavSim by Perez-Higueras et al. \cite{perezhunavsim}
MRPB 1.0 by Wen et al. \cite{mrpb} allows researchers to benchmark local planning approaches in diverse, complex, and dynamic environments within the Gazebo simulator. Other platforms for navigating in complex dynamic and cluttered environments were presented in \cite{xia2020interactive} ,\cite{portugal2019ros}, or \cite{gao2022evaluation}.
\\\noindent
Of note, the aforementioned platforms are either limited to static environments, have a very limited amount of navigation approaches, or worlds and scenarios to test, and/or require specific hard- and software and a tedious setup proceeding, which hampers their widespread usage. The BARN challenge by Xiao et al. \cite{xiao2022autonomous} aspires to compare state-of-the-art planners. The researchers provide an unified platform, which generates unknown environments to be mastered by the navigation approaches. The researchers also provide an unified API in which researchers can submit their navigation planners using an unified format. Although their initial challenge is based on static environments, their subsequent works, DynaBARN by Nair et al. \cite{nair2022dynabarn} extends this to dynamic environments. However, their platform is solely for the purpose of testing and also requires some non-intuitive setup steps in order to work properly. 
\\\noindent
In our previous works Arena-Bench \cite{kastner2022arena-bench}, we aspired to tackle the aforementioned issues by providing a holistic platform that provides all necessary modules to train, test, and evaluate navigation planners for both paradigms - learning-based and classic approaches. We provided a larger suite of robots and a randomized environment generator that can be used to generate an infinite amount of train and test environments. The training was possible via resource-efficient 2D simulations. We also provided several baseline classic planners to compare against in terms of a variety of evaluation metrics. However, a number of issues still remained that were recognizable based the continuous usage, user feedback, and Github issues. The proposed platform Arena-Rosnav 2.0 not only addresses all these issues but also adds a number of additional functionalities simplify usage and setup, and providing substantially more possibilities for extensions.

\begin{figure*}[]
    \centering
    \includegraphics[width=0.94\linewidth]{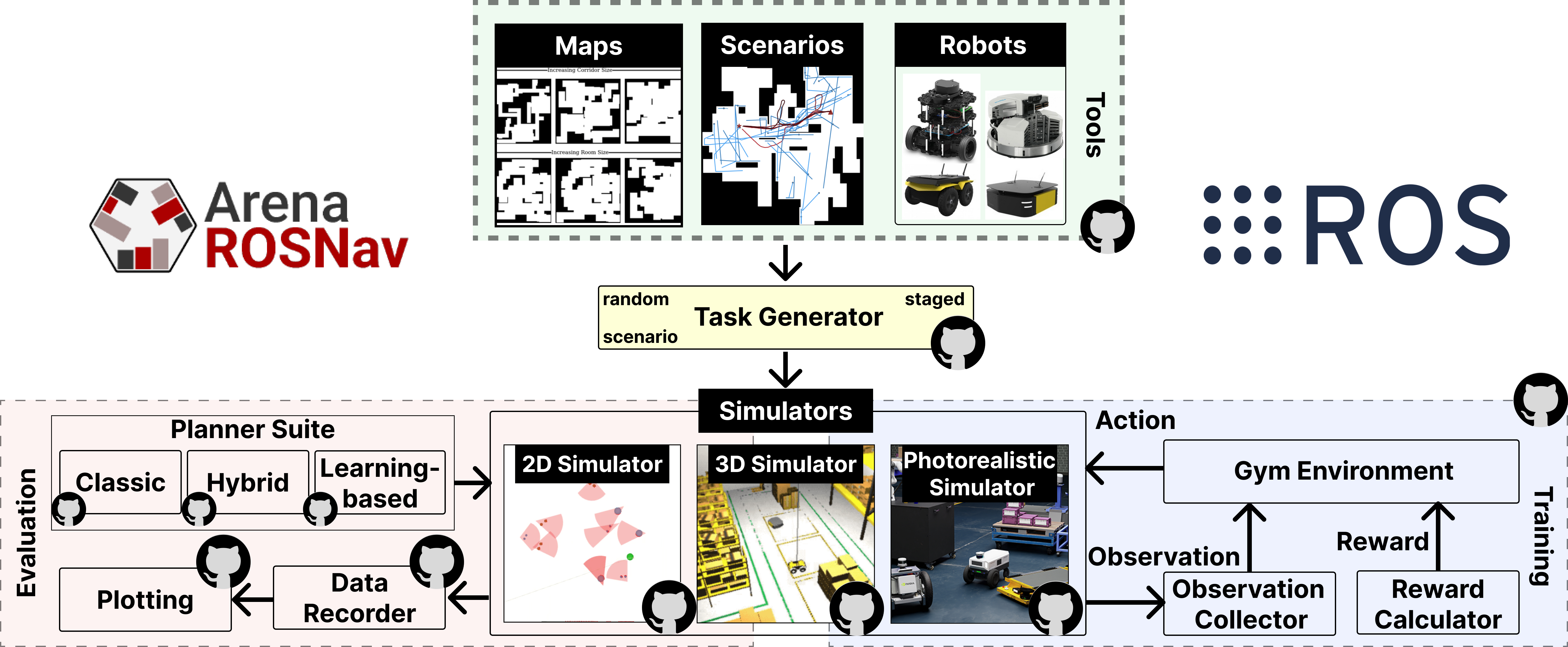}
    \caption{System Design of Arena-Rosnav 2.0. The system design was restructured fundamentally. It now contains the entities as seperate and independent github repositories, each of which can be utilized seperately and also integrated into other development pipelines. The task generator receives the task configurations in the form of maps, scenarios, and robots to instantiate the scenery in one of the available scenarios. The training can be conducted using the training pipeline, which is employing OpenAI gym environments and Stable Baselines 3 to provide the DRL algorithms. Benchmarking can be conducted with all planners within the planning suite. The evaluation can be conducted using the data recording and plotting class.}
    \label{fig:system}
\end{figure*}

\section{Methodology}
\noindent The main functionalities of our system can be divided into three categories: tools, training, and benchmarking. In the following, first, the system design and differences to the old version will be explained. Subsequently, each of those main modules will be described in more detail. 

\subsection{System Design and Differences to the First Version}
\begin{figure*}[!h]
    \centering
    \includegraphics[width=0.99\linewidth]{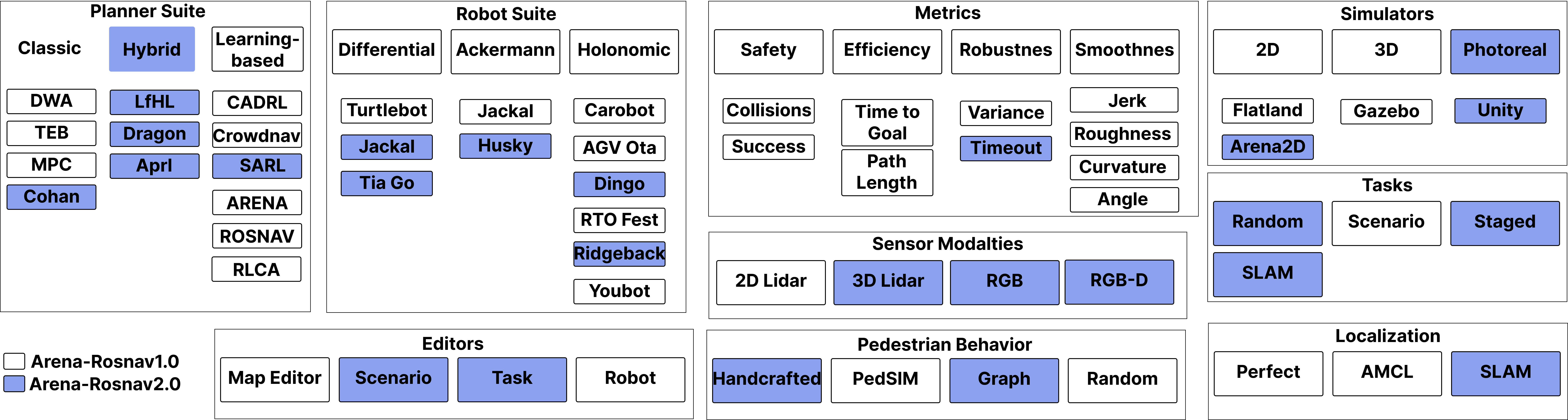}
    \caption{Differences between both versions. Blue are modules that were added in the second version of Arena-Rosnav. In particular, the hybrid planner class consisting of the majority of planners from the BARN challenge as well as other learning-based approaches were integrated into the planner suite. More robots are available and two new simulators were integrated using the provided API. Thus, more sensor modalities are now available such as RGB and 3D Lidar. Additionally, the pedestrian behavior can now be chosen using different behavior modules. Finally, SLAM as a localization method was added and new tasks were created.}
    \label{fig:differences}
\end{figure*}

\noindent Fig. \ref{fig:system} illustrates the system design of Arena-Rosnav 2.0. Fig. \ref{fig:differences} depicts the differences between the two version. Unlike the previous version, it is designed in a much more modular way with only a small number of mandatory modules to be installed. Each module is implemented inside a separate Github repository inside the Arena-Rosnav organization rather than within one single repository as in the first version. This leads to more flexibility in terms of usability and extensibility. Due to the this modularity and decoupled architecture of our platform, adding new simulators and planners is substantially simplified. To develop an unified and simple-to-use API, we analyzed a large number of repositories of state-of-the-art planners from other researchers, organizations, and challenges and provided an API that aspires to integrate these planners with as little adjustments as possible. In particular, we integrated all planners from the BARN leader-board of the 2022 BARN challenge \cite{xiao2022autonomous} as well as other state-of-the-art learning-based approaches such as CADRL \cite{everett2018motion}, CrowdNav \cite{chen2019crowd}, or the Cohan Planner \cite{singamaneni2021human}. The planners that were already present in the first version of our system, were outsourced to separate repositories.
\\\noindent
Another major improvement of this outsourcing is that not only new planners but also simulators can be integrated more easily. The core entity to facilitate the generation of worlds and tasks is the task generator. In the first version, this task generator was designed specifically to work with the existing simulators Flatland and Gazebo to spawn and delete obstacles and robots and generate worlds. In Arena-Rosnav 2.0 the task generator is an independent entity, which provides an API with the objective to integrate new simulators as simple as possible. Therefore, the user has to specify four main function calls for spawning or deleting objects. In the process, we integrated the Unity simulator and the Arena2D simulator of our previous works \cite{kastner2020deep}. 
\noindent To utilize these APIs and improve user understanding, we provide a profound documentation of the system under \href{https://arena-rosnav.readthedocs.io/en/latest/}{\color{blue}https://arena-rosnav.readthedocs.io/en/latest/}.
\\\noindent
Another addition is the option to deploy multiple robots and independent navigation stacks simultaneously. This enables development of Multi-agent-reinforcement-learning (MARL), which was not possible within the first version (see the supplementary video for demonstrations of MARL agents developed on this platform). 
\\\noindent
Finally, the installation procedure is significantly simplified and can be conducted with a single command. We also reworked the evaluation class, which now includes more options for customization and more metrics. The user has the option to customize plots and is not dependent on a complicated configuration file anymore. The task mode was extended to also include three more tasks: when the exact position of the robot is not known, when the map is unknown or when both are unknown. This is in accordance with the BARN challenge, which might be helpful for outdoor navigation in unknown environments. 
\begin{table}[htbp]
\centering
\setlength{\tabcolsep}{12pt}
\renewcommand{\arraystretch}{1}
\setlength{\tabcolsep}{.5\tabcolsep}
\footnotesize
\caption{Available Navigation Planners}
\begin{tabular}{@{}lccccc@{}}\toprule
Classic & Hybrid & Learning-based   \\ 
\midrule
TEB \cite{rosmann2015timed} &  Applr \cite{xiao2020appld} & ROSNavRL \cite{kastner2021towards}  \\
DWA \cite{khatib1986real} &  LfLH \cite{xiao2022motion} & RLCA \cite{long2018towards}  \\
MPC \cite{rosmann2019time} &  Dragon \cite{xiao2022autonomous} & Crowdnav \cite{chen2019crowd}  \\
Cohan \cite{singamaneni2021human} &  TRAIL \cite{xiao2022autonomous} & SARL \cite{li2019sarl}  \\
 &   & Arena \cite{kastner2020deep}  \\
  &   & CADRL \cite{everett2018motion}  \\
    &   & Navrep \cite{dugas2020navrep}  \\
\bottomrule
\end{tabular}
\label{tab:planners}
\end{table}
\noindent Alongside that aspect, we also added SLAM approaches as a different and more realistic localization approach to the perfect localization and AMCL, which can be chosen by the user. The editors have a reworked user interface for easy and fast configuration and accessibility. To make the behavior of pedestrians more realistic, we also integrated Pedsim and the Graph behavior algorithm of Tsoi et al. \cite{tsoi2022sean}. The training and testing can be conducted cross-compatible across all simulation environments due to the unified task manager.

\subsection{Arena-Tools}
\noindent
Arena-Tools provides user interfaces that allows for generation of maps, scenarios, robots, and tasks. It is implemented as a Python application, which the user can download and use out-of-the-box. The user is provided an user interface in which the necessary parameters can be given as input. The resulting configuration files are designed to be read out by the other entities such as the task manager or the training pipeline. We have reshaped and redesigned the interface of Arena-Tools to be more intuitive. Furthermore, the scenario editor now contains more possibilities to design scenarios such as more behavior modes. This gives the user more flexible control over the design of the scenarios for training and testing. Arena-Tools is available under \href{https://github.com/Arena-Rosnav/arena-tools}{\color{blue}https://github.com/Arena-Rosnav/arena-tools}. For a more profound description and visualization of Arena-Tools, we refer to our previous works \cite{kastner2022arena-bench}.

\subsection{Training}
\noindent For training DRL agents, we utilize the training pipeline similar to the one proposed in our previous works. In particular, the OpenAI Gym interface is utilized, which provides an interface to abstract the DRL aspect from the simulator. Stable baselines 3 \cite{raffin2019stable} is used to include a number of DRL algorithms. The Observation collector receives the input from the simulator and sends it to the gym environments. The reward calculator defines the specific reward systems. We employ a basic reward system that should be utilized as a high number of DRL agents were already successfully trained with it. Additionally, this basic reward system can be fine-tuned and extended using more customized rewards e.g. to teach the agent more efficient but aggressive behavior or vice versa: to train a considerate agent. Due to the different robots, in- and output sizes are to be adjusted accordingly. It is recommended to utilize the 2D environment for training for a resource-efficient training - compared to the 3D simulators. However, we also made training in Gazebo and Unity possible. Therefore, additional sensors were integrated, more specifically RGB and RGB-D. However, due to the resource-intensive training, only a validation of function was conducted and a profound training is yet to be released when more powerful hardware is available.

\subsection{Benchmarking}
Benchmarking different navigation planners is an important functionality of our system. Therefore, a number of modules are integrated . In particular, the data recording module to record necessary information, the preprocessing module, as well as the plotting module. For the latter, evaluation metrics play an important role.
\subsubsection{The Data Recording Module}
\label{sec:data-recording}
The data produced by the deployed approaches must be recorded properly. Thus, an important new module of the system is the data recorder. In the previous version, Rosbag files were used. However, due to a number of limitations, we re-implemented the recording pipeline to include more functionalities and make the evaluation step more customizable. Since the modules of this work rely on ROS as standard middleware, ROS standard messages will be recorded. More specifically, the laser scan pattern, and robot odometry will be recorded since these two topics provide the necessary information to compute the large number of metrics. Laser scan messages will be used to calculate collision rates and clearing distances, while the robot odometry provides information about the robot path. Within ROS, these messages are equipped with a time stamp which allows calculation of metrics such as the time required to reach the goal. Additional parameters, like the obstacle movements or world parameters, like the map are also recorded to provide more information. The data recorder is conceptualized in a modular way to be integrated in any given ROS related project.

\subsubsection{The Plotting Module}
Meaningful visualization of the gathered information could improve understanding significantly and aid in a fast performance assessment. Therefore, the plotting class is provided, which can produce unified plots to make comparison and assessment of the results more intuitive and simple. A number of plotting options are provided inside the plotting class to visualize relevant information expressively.
\label{sec:plotting}

\subsubsection{Metrics}
Evaluation metrics play a significant role to assess performance of any given system. Therefore, a variety of metrics are provided inside the our evaluation class. Four main aspects to assess the performance of the planners are defined: navigational safety, efficiency, robustness, and smoothness, which can be assessed using measurements acquired during evaluation runs. 
\begin{table}[!h]
\centering
	\setlength{\tabcolsep}{0.2pt}
	\renewcommand{\arraystretch}{0.5}
		\caption{Evaluation metrics}
	\begin{tabular}{l l r}
		\hline
		Metric  &Unit & Explanation    \\ \hline
		Success Rate$^{2}$ & \%        & Runs with < 2 collisions          \\ 
		Collision$^{1,2}$ & -        & Total number of collisions\\
		Time to reach goal$^{2}$& [$s$] & Time required to reach the goal   \\ 
		Path Length$^{1,2}$  & [$m$]    & Path length in m           \\ 
		Velocity (avg.)$^{2}$     & [$\frac{m}{s}$]  & Velocity of the robot \\
		Acceleration (avg.)$^{2}$ & [$\frac{m}{s^2}$] & Acceleration of the robot \\
		Movement Jerk$^{2}$ & [$\frac{m}{s^3}$] & Derivation of Acceleration \\ 
		Curvature(avg.,max.,min.,norm.)$^{2}$ & [$m$] &  Degree of trajectory changes\\
		Angle over length$^{2}$  & $[\frac{rad}{m}]$ & Curvature over the  path-length \\
		Roughness$^{2}$ & -                 & Quantifies trajectory smoothness \\ 
		Clearing Dist.(avg.,max.,min.,norm.)$^{2}$ & $[m]$ & Distance kept to obstacles \\
		\hline
	\end{tabular}
    \footnotesize{$^1$Quantitative metric, $^2$Qualitative metric}\\
	\label{tab:metrics}
	
\end{table}
\\\noindent The most intuitive metric aspects are safety and efficiency, which are both paramount for real-world usage. Safety indicates that the planners can reach the destination without collisions or risky maneuvers potentially causing danger to other actors within the environment. It can be assessed by measuring the collision rates produced by the agent over a given number of runs. Another aspect to quantify safety is the clearance distance. A small clearance distance may indicate risky behavior if the robot is driving too close to obstacles, which might be more efficient yet dangerous. For this reason, both metrics, safety and efficiency often times counteract with one another. At the same time, a high efficiency is important to save time and costs. Reasonable indicators for the efficiency of an approach are the length of the path and the time it takes to reach the destination over a given number of episodes.
\noindent
\begin{figure*}[!h]
	\centering
	\includegraphics[width=0.93\textwidth]{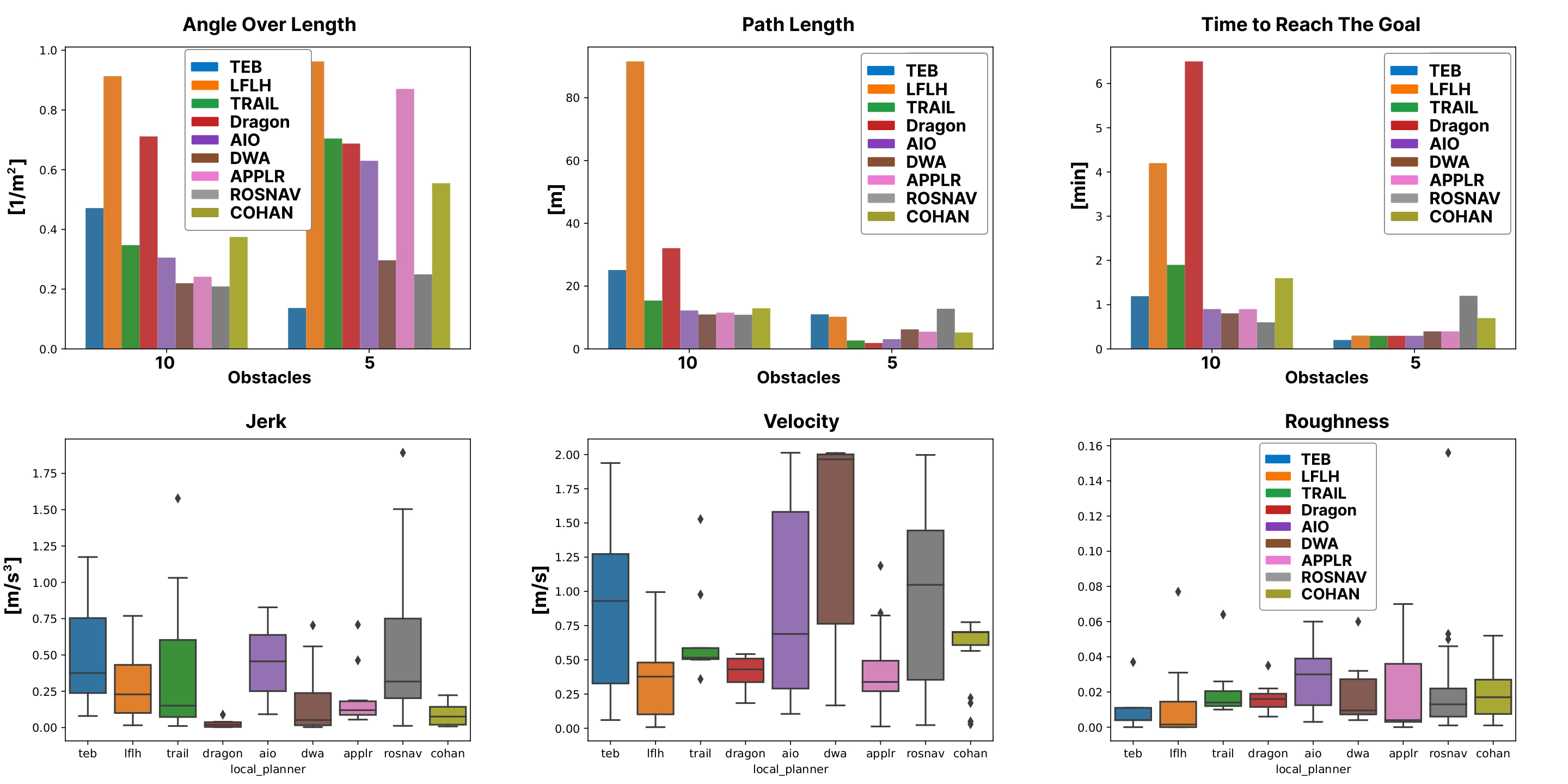}
	\caption{Quantitative evaluations of all planners on a map with 5 and 10 randomly generated obstacles. Each planner was run for 30 times.}
	\label{fig:quanti}
\end{figure*}
Further, the robustness of the system is a paramount factor to assess the systems' performance. A high robustness indicates a certain performance guarantee over a large number of runs. Robustness can be measured by evaluating success rates over a series of test runs or by observing the occurrences of time-outs. Additionally, the variance of the results, i.e., whether and how often there are significant outliers between test runs, provides an important indicator for the reproducibility of the results and as a consequence, for the robustness of the approach. A large variance would indicate that the results are not reproducible and each test run leads to new results.
\\\\\noindent
Finally, navigational smoothness is another important aspect to consider. Smoothness indicates how smoothly the agent moves along its path or if there are many unnatural jerks in motion. It also indicates how many turns the trajectory has and if there are many detours, which directly correlates with the efficiency metrics. 
A high smoothness is often aspired as it not only saves computational power but also constitutes to more natural behavior of the agent.

\section{Evaluation}

\subsection{User Feedback about Arena-Rosnav 2.0}
\noindent 
A major focus of Arena-Rosnav 2.0 was to improve availability and lower the barrier for entrance as the previous version some user reported problems in the installation processor training and evaluation due to a number of issues such as wrong versions, lack of documentation, hardware, and software bugs. These problems were all solved in the new version and the structure was re designed from ground up to simplify the installation process as only the necessary and desired modules will be downloaded and installed. Due to the restructuring of the platform, a number of improvements in terms of installation processes could be accomplished. Previously, an installation of the first version took on a standard development PC, approximately 10 min. To install the same functionalities as the first version, using the new platform requires less than 2 minutes using the same development PC. Further, due to the restructuring, the size of the repository is substantially reduced. The results are listed in Table \ref{tab:specs}.
\begin{table}[H]
    \centering
    \begin{tabular}{lll}
        \toprule
        \textbf{Component} & \textbf{Setup Time} & \textbf{Size} \\
        \midrule
        Arena-Rosnav & 21.5 min & 5.7GB \\
        Arena-Rosnav 2.0 & 4.2 min & 0.6GB \\
        \bottomrule
    \end{tabular}
    \caption{Setup comparison}
    \label{tab:specs}
\end{table}
\noindent
To evaluate the user experience, we asked 8 researchers from different universities across Europe and one researcher from a US university to install and test out our system. The users installed and tested out the major functionalities of the system including training, evaluation, and plotting. Subsequently, they provided written feedback.
\\\noindent
The majority of users noted the easy installation process and good documentation. Three of the users were also familiar with the first version and pointed out the clear improvement and cleanliness of the code base which allows for customized installation of only required modules rather than the mandatory installation of one whole repository, which took significantly more time for the old version. In contrast, the new version's size at its core is less than 600 MB, compared to over 5.7 GB for the old version, which poses an almost ten-fold compression. Six of the 9 users were roboticists and praised the availability of a large number of robot models. They also noted that the robots made realistic movements in all simulators even the 2D one. However, five of the 10 people noted that the random mode for pedestrians resulted in unrealistic movements and at times buggy situations when multiple persons crossed paths or were stuck in a small corner or alley. This issue is solved when using either the handcrafted or the graph mode. Seven users commented on the helpfulness of the plotting class, which is now more customizable and does not require to input a large number of parameters like in the previous version.
Two of the testers had problems running the newly integrated hybrid planners due to conflicting package versions. We made sure to resolve that problem and point specific requirements more clearly in our documentations.
\\\noindent
Other researchers noted the high number of state-of-the-art navigation planners that are present in the planner suite. One researcher plans to utilize our platform to develop a planner for the 2023 BARN challenge. We are currently in talks with the BARN organizers for potential synergies of our two similar systems such that Arena-Rosnav 2.0 could be used to serve as a platform for conducting the challenge in the future. We also paid attention to a number of Github issues regarding the training and installation process, which sometimes required specified dependency versions. Using these hints and comments, we incorporated all user feedback into our new system, which resulted in a much improved version as recognized by the testers.

\begin{figure*}[!h]
	\centering
	\includegraphics[width=\textwidth]{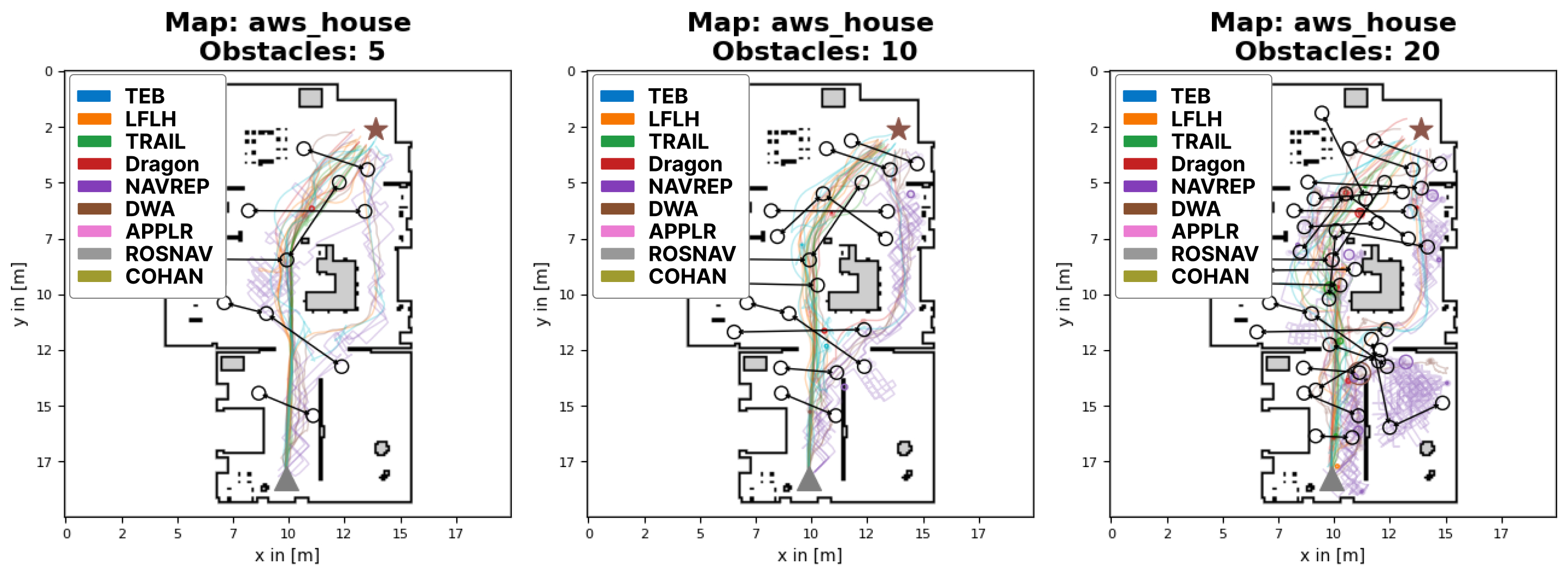}
	\caption{Qualitative results of all planners on the small warehouse map of the AWS Robomaker project. White circles and their respective black paths represent dynamic pedestrian movements. Red circles indicate collisions.}
	\label{fig:quali}
\end{figure*}

\subsection{Quantitative Evaluations of Navigational Performance}
\noindent One major aspect to be done with our platform is the extensive benchmarking of the navigation approaches. The metrics listed in Table \ref{tab:metrics} can be plotted in a number of ways. We provide a template within our evaluation class to plot different metrics and results using a number of options.
To demonstrate the capabilities of our platform, we conducted a quantitative evaluation on all planners. The planners were tested on a specific map with 5 and 10 pedestrians moving using the Pedsim social force model \cite{helbing1995social}. There are a variety of navigational aspects such as safety, efficiency, robustness, and smoothness that can be evaluated depending on the use case. An extensive evaluation in terms of safety and robustness of several planners were conducted in our previous version Arena-Bench \cite{kastner2022arena-bench}. In this work, to demonstrate the evaluation capabilities, we focus on navigational smoothness and efficiency. Thereby, the efficiency is indicated by the path length and time to reach the goal. The movement jerk expresses the rate at which the robot changes its acceleration with respect to time. Other indicators of smoothness include the angle over the length of the total trajectory, and the roughness of the trajectory. Although our platform enables deployment of all planners on several different robots, in this work, we deployed all planners only on the Clearapath Jackal robot for exemplary demonstration. The results are illustrated in Fig. \ref{fig:quanti}.
\\
\noindent \textbf{Navigational Efficiency:}
With regards to navigation efficiency, it is observed that the hybrid planners Dragon and LFHL performs worst with an average of over 6.3 and 4 minutes to reach the goal respectively. However, the path length for Dragon is significantly smaller with an average of 30 m compared to over 90 m for LFLH, which indicates that the Dragon planner got stuck more often, whereas the LFLH might produced more roundabout paths to reach the goal. This however, is only observed in the scenario with 10 obstacles. In the scenario with 5 obstacles, all planners accomplish competitive results with less than 10 m and 1 min to reach the goal. Thereby, all planners except for Rosnav and Cohan attain similar results with the Rosnav and the Cohan planner requiring over 1 and 0.8 min respectively.
\\
\noindent \textbf{Navigational Smoothness:} In terms of navigational smoothness, it is observed that the hybrid planners APPLR, Dragon, and Trail attain the best results with a jerk of under 0.2 $m/s^3$. Surprisingly, the classic DWA planner achieves competitive performance although employing a stop-and-backwards movements once obstacles are perceived. The planners Cohan and LFLH also achieve competitive results with 0.2 and 0.25 $m/s^3$ respectively. Our Rosnav planner and the classic TEB planner produce mediocre results with a higher jerk compared to the other planners except for the AIO planner, which is caused by jerky movements or the typically stop-and-backwards movements of TEB once an obstacle is perceived.
The AIO planner performs worst with an average jerk of almost 0.5 $m/s^3$. This is expected since the AIO planner is a combination of the Rosnav and TEB planner both of which produce the highest jerk.
The findings indicate that the hybrid planners accomplish the best performance in terms of navigational smoothness followed-up by the classic approaches and learning-based approaches producing the worst performance.

\subsection{Qualitative Evaluations of Navigational Performance}
\noindent Using our platform, qualitative evaluations are also possible. Unlike the quantitative evaluations, where randomly generated episodes can be used and thousands of episodes run, qualitative evaluations should visualize the trajectory. Therefore, to ensure consistent situations for all planners, the scenario mode should be used where the user can define the exact pedestrian behaviors. Therefore, we utilized the AWS warehouse world of the AWS Robomaker project. We defined the pedestrian movements using the handcrafted option in Arena-Tools to define the exact pedestrian movements for the scenario. Subsequently, we run every planner for a total of five times. The results are plotted in Fig. \ref{fig:quali}.
It is observed that most of the planners reach the goal in a straightforward manner. The planner Navrep and TEB however produce a high number of roundabout paths. For Navrep the researchers already pointed out that the planner produces jerky motions, which can be clearly visualized within these plots. In scenarios with a higher number of obstacles, it is observed that the other planners also create roundabout paths to reach the goal. In the scenario with 20 obstacles almost all planners have difficulties reaching the goal and a high number of collisions occurred (marked in red). The qualitative evaluations can provide a visualization of the trajectory of the planners to give an insight in the behavior. It can also visualize where collisions occurred to spot dangerous areas. However, using a high number of planners within one plot can result in complex and incomprehensible plots. It is therefore recommended to plot no more than four planners at once.

\section{Conclusion}
\noindent In this paper, we proposed Arena-Rosnav 2.0 an improved version of our previous work Arena-Bench and Arena-Rosnav. It employs a totally new structure, which was designed to include every module as a separate and independent Github repository rather than a composite repository as was the case with the first version. This way, the installation process is substantially simplified and can be conducted within a single command. Additionally, extensibliity and modulartiy is ensured, which simplifies adding more features such as planner or simulators. Therefore, we provide clear and easy-to-use APIS, which are well documented within our profound documentation, which also include step-by-step tutorials of the most important aspects. To exemplary showcase the platforms functionality we integrated two new simulators and several state-of-the art navigation approaches into our new platform. The platform contains all common robot kincematics, planning types as well as realistic localization an mapping approaches. We extended the task modes and also made the simulation and the behavior of obstacles more realistic. Further, we reworked the evaluation class to provide more metrics and allowing more customizable plotting. The new platform was evaluated by first, conducting a user study, which provided mostly positive feedback, and secondly, by testing all planners on several robots qualitatively and quantitatively. Future works include the incorporation of more simulators and benchmarking of these in terms of training capabilities or comparing the different sensor modalities. Furthermore, we aspire to include a real robot automation benchmarking pipeline into our system and provide a web-based user interface to simplify usage of our platform even further. In the future, our platform could also serve as a platform to conduct challenges and benchmarks and we are currently in talks with a number of other research groups for potential synergies.

%\section*{Acknowledgement}
%We acknowledge help with the production of the spot welds and with the chisel test by Hubert Suwala.

\addtolength{\textheight}{-1cm} 

%%%%%%%%%%%%%%%%%%%%%%%%%%%%%%%%%%%%%%%%%%%%%%%%%%%%%%%%%%%%%%%%%%%%%%%%%%%%%%%%

%%%%%%%%%%%%%%%%%%%%%%%%%%%%%%%%%%%%%%%%%%%%%%%%%%%%%%%%%%%%%%%%%%%%%%%%%%%%%%%%

%%%%%%%%%%%%%%%%%%%%%%%%%%%%%%%%%%%%%%%%%%%%%%%%%%%%%%%%%%%%%%%%%%%%%%%%%%%%%%%%

%%%%%%%%%%%%%%%%%%%%%%%%%%%%%%%%%%%%%%%%%%%%%%%%%%%%%%%%%%%%%%%%%%%%%%%%%%%%%%%%
\typeout{}
\bibliographystyle{IEEEtran}
\bibliography{main}

\end{document}